\documentclass{article}
\usepackage{PRIMEarxiv}
\usepackage[utf8]{inputenc} 
\usepackage[T1]{fontenc}    
\usepackage{hyperref}       
\usepackage{url}            
\usepackage{booktabs}       
\usepackage{amsfonts}       
\usepackage{nicefrac}       
\usepackage{microtype}      
\usepackage{lipsum}
\usepackage{fancyhdr}       
\usepackage{graphicx}       
\graphicspath{{fig/}}     
\usepackage{xcolor}
\usepackage{subcaption}

\usepackage[
backend=biber,
style=authoryear,
uniquename=false
]{biblatex}
\bibliography{ref.bib}

\usepackage{booktabs}

\usepackage[title]{appendix}
  
\title{
Transformer Encoder for Social Science
\thanks{This project is funded by Niehaus Center for Globalization and Governance. We are grateful for the advice from David Turner, Marc Ratkovic, Walter Mebane, Sirus Bouchat, Brian Klobucher, and Pradhan Abhijeet.}
\thanks{GitHub:\href{https://github.com/haosenge/TESS}{\url{https://github.com/haosenge/TESS}}, Hugging Face: \href{https://huggingface.co/hsge/TESS_768_v1}{\url{https://huggingface.co/hsge/TESS_768_v1}}}
}

\author{
  Haosen Ge \\
  Princeton University\\
  hge@princeton.edu\\
   \And
  In Young Park \\
  Boston University\\
  ip@bu.edu\\
     \And
  Xuancheng Qian \\
  Princeton University\\
  xuancheng.qian@princeton.edu\\
     \And
  Grace Zeng \\
  Princeton University\\
  gracezeng@princeton.edu\\
}

\begin{document}
\maketitle

\begin{abstract}
High-quality text data has become an important data source for social scientists. We have witnessed the success of pretrained deep neural network models, such as BERT and RoBERTa, in recent social science research. In this paper, we propose a compact pretrained deep neural network, \textit{Transformer Encoder for Social Science (TESS)}, explicitly designed to tackle text processing tasks in social science research. Using two validation tests, we demonstrate that TESS outperforms BERT and RoBERTa by 16.7\% on average when the number of training samples is limited (<1,000 training instances). The results display the superiority of TESS over BERT and RoBERTa on social science text processing tasks. Lastly, we discuss the limitation of our model and present advice for future researchers.
\end{abstract}

\section{Introduction}

Text data has become a central piece of social science research. In particular, political scientists make use of text data such as legislative or legal documents \autocite{truex2020authoritarian,chen2021settling}, news articles \autocite{king2003automated, huffkertzer2017, chang2019, wasow2020, kim2021political,halterman2021corpus, barbera2021}, speeches \autocite{monroe2008fightin,enamorado2021scaling, boussalis2021, goplerudsmith2022}, and open-ended survey questions \autocite{roberts2014structural, mueller2022, manekinmitts2022}. 

We have witnessed the success of applying pretrained DNNs, such as BERT and RoBERTa \autocite{devlin2018bert,liu2019roberta}, on a wide array of text-related supervised learning tasks \autocite{kim2021learning,wankmuller2021neural}. However, while DNNs can achieve near-human performance, training DNNs well usually requires a significant amount of data, and annotating a large training set is often costly in social science research settings. 

To reduce the cost of using DNNs in social science, we propose a pretrained DNN model designed for social science text processing tasks, Transformer Encoder for Social Science (TESS). Our model offers four advantages over the popular models used by political scientists (e.g., BERT and RoBERTa) on social science-related tasks:
\begin{itemize}
    \item \textbf{Less Data Hungry:} 
    TESS demonstrates superior performance in small training sample settings compared to other popular models.
    \item \textbf{Smaller Size:} TESS is about half the size of BERT and RoBERTa, which makes it easy to use on machines with limited memory.
    \item \textbf{Longer Input Sequence}: TESS can read text sequences up to 768 tokens at a time, while most models on the market take text sequences up to 512 tokens as input.
    \item \textbf{Better Performance:} TESS shows a modest increase in performance than other models on social science related tasks.
\end{itemize}
Therefore, we believe TESS can be a good alternative to many pretrained DNNs when conducting social science research. Like BERT and RoBERTa, TESS adopts the structure of a transformer encoder \autocite{vaswani2017attention}. The encoder structure uses parameters more efficiently than the decoder (e.g., GPT-2) but cannot be used for language generation purposes. Since most language processing tasks encountered by social scientists do not include natural language generation, we believe that the encoder structure fits researchers' needs better than the decoder structure.

However, we improve upon BERT and RoBERTa by introducing the following changes:
\begin{itemize}
    \item \textbf{Embedding Dimensions and Weight Sharing:} Following the findings by \textcite{lan2019albert}, we reduce the wording embedding dimension from 768 to 128 and enable weight-sharing. The model shares the attention weights across all layers but uses individual feed-forward networks. This architecture significantly reduces the number of required parameters (by around 50\%) while still preserving the performance.
    \item \textbf{Pretraining on a Social Science Corpus:} We conducted masked language modeling on a considerable social science corpus that includes U.S. congressional bills, annual reports of U.S. public firms, Preferential Trade Agreements, UNGA Resolutions and U.S. Court Opinions.
    \item \textbf{Increasing Input Sequence Length:} we set the input sequence length to be 768 instead of 512, as most text processing tasks in social science require dealing with long documents.
\end{itemize}

To evaluate the model's performance, we conducted two validation studies with tasks closely related to political science research. First, we test the model's performance using the dataset published by \textcite{kim2021learning}. In their project, they propose a multi-class classification task in which human annotators are tasked with comparing the similarity of two subsections of U.S. congressional bills. Later, they test the performance of a series of models, including RoBERTa and BERT. In the second validation test, we utilize the original data proposed by \textcite{ge2021measuring}. He proposed a binary sequence classification task to distinguish whether a text sequence contains information about regulatory barriers in certain countries. Our model, TESS, achieves a 6.6\% increase in full sample performance compared with BERT and RoBERTa. Furthermore, we test the model's performance under the small training sample scenario. On average, TESS outperforms BERT and RoBERTa by 16.7\% when the number of training samples is below 1,000.

In the next section, we will introduce the model, corpus, and pretraining tasks. The final section will present results from validation analyses.

\section{TESS: Transformer Encoder for Social Science}

\subsection{Training Corpora}

We collected over 35GB of compressed text from multiple sources. \href{copora}{Table \ref{copora}} offers a brief view of the text corpora.
\begin{table}[h!]
\centering
\caption{Corpora}
\begin{tabular}{ll}
\toprule
\textbf{TEXT}                 & \textbf{SOURCE}              \\
\toprule
Preferential Trade Agreements & Texts of Trade Agreements    \\
Congressional Bills           & \textcite{kornilova2019billsum}                      \\
UNGA Resolutions                &  UN                            \\
Firms' Annual Reports         & \textcite{loughran2016textual} \\
U.S. Court Opinions           & Caselaw Access Project    \\
\bottomrule
\end{tabular}
\label{copora}
\end{table}

The corpora differ significantly from the Wikipedia corpus, which is commonly used in pretrained language models. They are, however, close to the text that many political scientists encounter in research. Recent studies show that pretraining on different corpora will significantly increase the model's performance in the corpora \autocite{beltagy2019scibert,gururangan2020don}. We cannot collect comprehensive data from news outlets or twitter due to copyright issues. 

\subsection{Model Pretraining}

Similar to BERT and RoBERTa, TESS is a 12-layer transformer encoder. However, all 12 attention layers share the same weights, but the feed-forward layer has its individual weights. This architectural design is based on the results from \textcite{lan2019albert}, which shows that sharing attention weights do not negatively affect model performance while sharing the feed-forward network weights does. By sharing attention weights, we reduce the model size by 50\% but still preserve the model performance.

We use the classic masked language modeling to pretrain the encoder. Following the literature, the masking probability is set to be 15\%. However, we allow n-gram masking, which is shown to lead to superior performance. In the original BERT training, the authors also implement next sentence prediction. We exclude the next sentence prediction task because our text is usually comprised of multiple sections that are not closely connected. Our experiment also found that training with next sentence prediction leads to worse performance than only training with masked language modeling.

We train the model for 120K steps in batches of 2,000 to facilitate model convergence. We use the AdamW optimizer with a learning rate of 1e-4. The model is trained on 4 NVIDIA A100 GPUs hosted by Lambda Labs GPU Cloud.

\section{Validation}

\subsection{Bill Similarity}

\textcite{kim2021learning} propose a five-class classification task to compare bill similarity. The similarity between two subsections of bills ranges from 0 (unrelated) to 4 (identical). The authors manually annotated 4,721 pairs of bill subsections. Among all pairs of subsections, the unrelated bills account for more than 50\% of total examples, which shows considerable data imbalance. In the paper, the authors find that the best performance is achieved by RoBERTa (F1 = 76.8).

We fine-tuned TESS on the training set with 3,305 human-annotated samples and tested the model performance on the authors' validation set. We experimented with learning rates = \{1e-5, 3e-5, 5e-5\} and epochs = \{1, 3, 5\} and report best outcomes for each model. 

\begin{figure}[h!]
\centering
\begin{subfigure}{.5\textwidth}
  \centering
  \includegraphics[width=\linewidth]{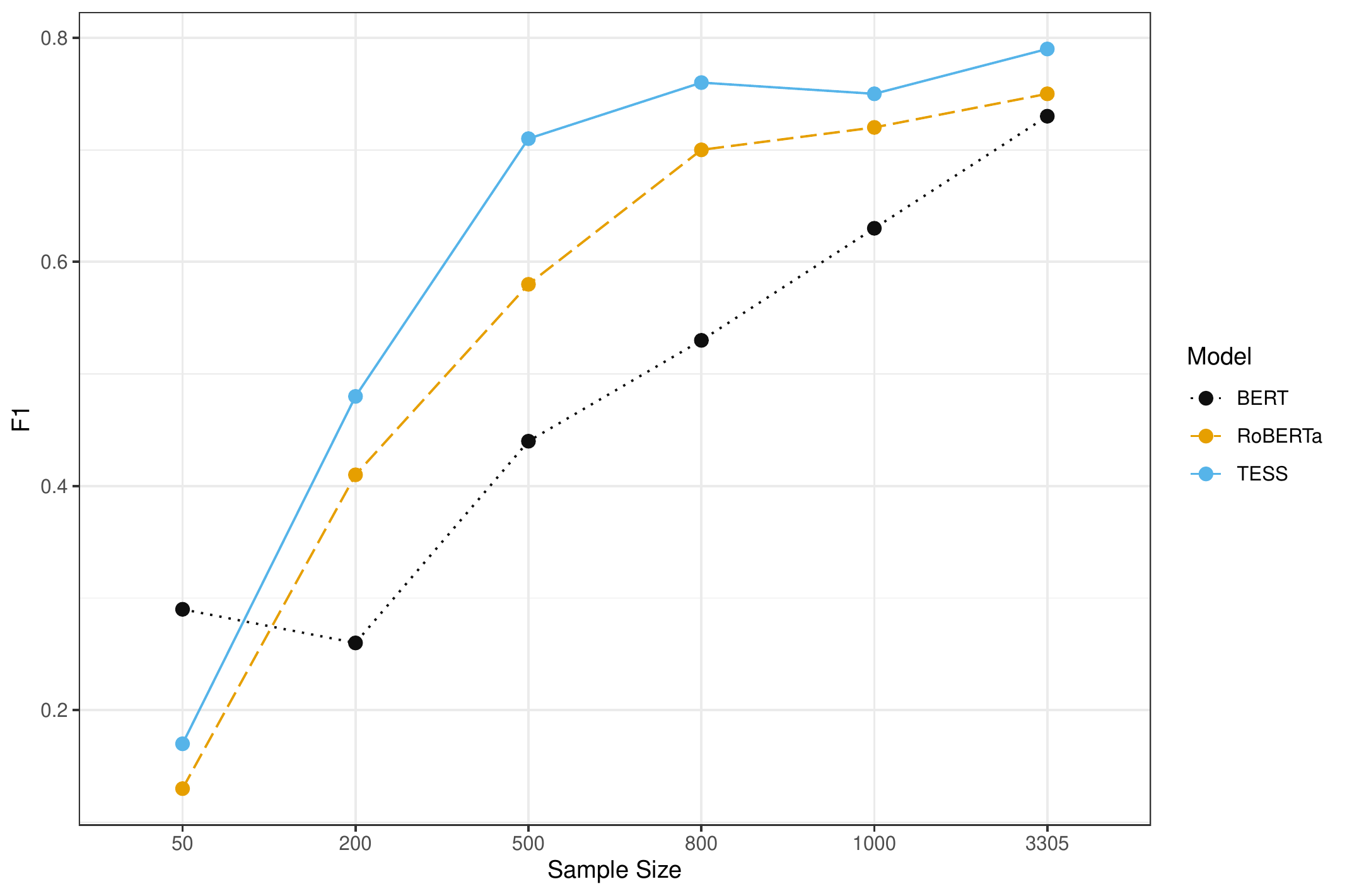}
  \caption{Bill Similarity}
  \label{fig:sub1}
\end{subfigure}%
\begin{subfigure}{.5\textwidth}
  \centering
  \includegraphics[width=\linewidth]{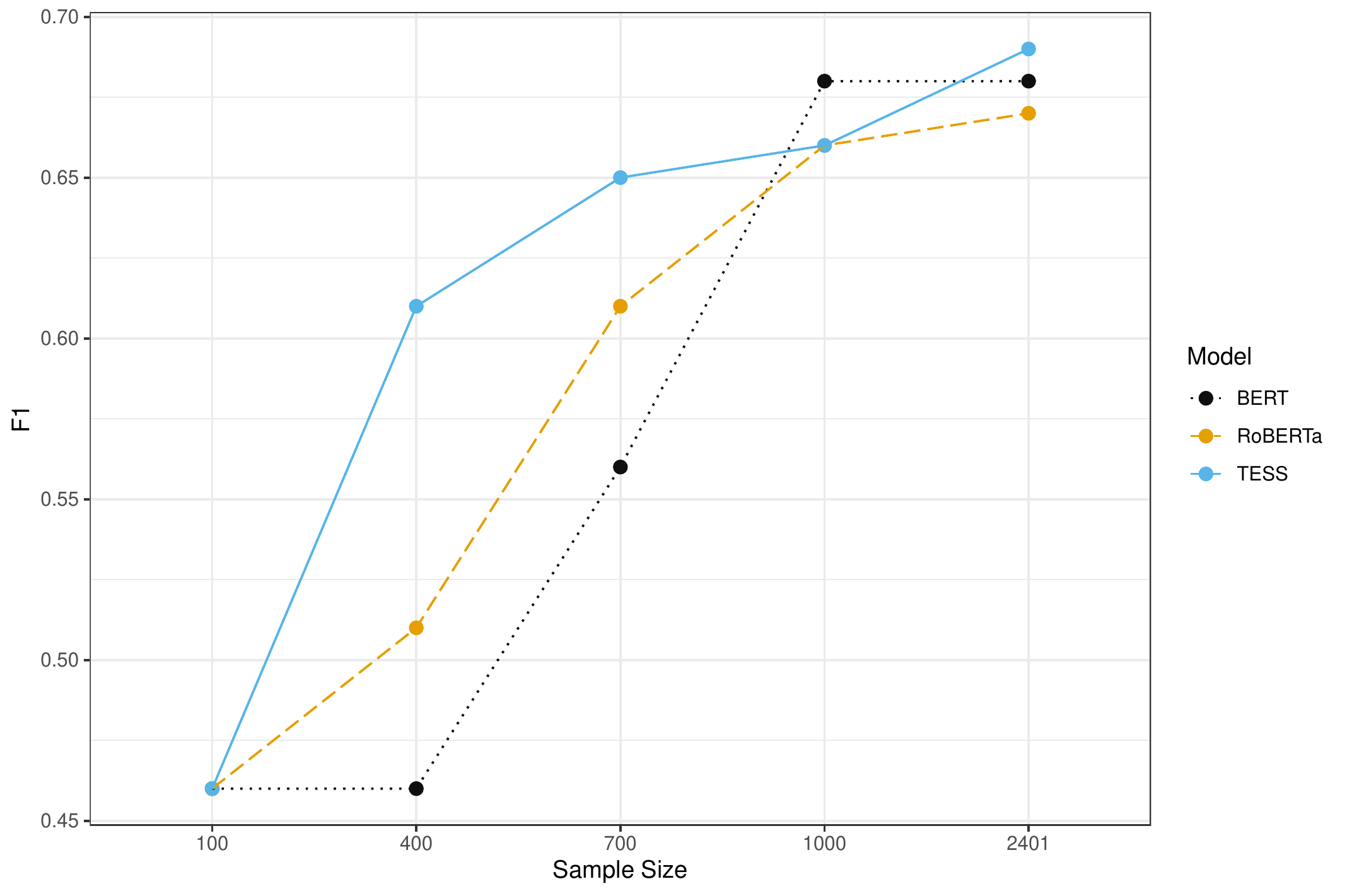}
  \caption{Regulatory Barrier}
  \label{fig:sub2}
\end{subfigure}
\caption{Validation Results}
\end{figure}

In \href{fig:sub1}{Figure \ref{fig:sub1}}, we present the comparison of TESS, RoBERTa, and BERT. To test model performance under small sample scenarios, we fine-tune each model using the sample sizes \{50, 200, 500, 800, 1000, 3305 (full sample)\}. It can be observed that TESS outperforms BERT and RoBERTa across almost all sample sizes except for the sample size of 50. When trained on the full sample, TESS achieved an F1 of 0.79, higher than BERT, RoBERTa, and the authors' best results in the paper. With only $500$ training instances ($1/6$ of the total training samples), TESS can achieve an F1 above 0.7, which showcases the model's ability to reduce researchers' annotation costs.

\subsection{Regulatory Barrier}

In the second validation study, we utilize the original data set compiled by \autocite{ge2021measuring}. They manually annotate sentences and paragraphs from U.S. public companies' annual reports (i.e., 10-Ks). By reading the sentences, they aim to classify whether a sentence or section contains information about the regulatory barrier encountered by the firm. We use three instances from the training set to illustrate the task:
\begin{itemize}
    \item \textbf{Positive:} We have difﬁculty gaining market share in countries such as Japan because of regulatory restrictions and customer preferences.
    \item \textbf{Positive:} Burdens of complying with a variety of foreign laws, including more protective employment laws affecting our sizable workforce in Germany.
    \item \textbf{Negative:} We are obligated to comply with various foreign, federal, state, and other environmental laws and regulations, including the environmental laws and regulations of the United States, Iceland, China, and the EU.  
\end{itemize}

We again experiment with learning rates = \{1e-5, 3e-5, 5e-5\} and epochs = \{1, 3, 5\}. When trained on the full training sample ($n$ = 2,401), TESS has the best performance among the three models (F1 = 0.69). We further investigate the small sample scenarios, $n = \{100, 400, 700, 1000\}$. Similarly, TESS outperforms BERT and RoBERTa in small training samples: the gap between TESS and the other two models shrinks when the sample size increases.

The two validation results prove TESS' effectiveness in saving annotation costs without sacrificing the large sample performance.

\section{Conclusions and Limitations}

This paper proposes a transformer encoder  designed for the social science research context. By validating on political science text processing tasks, we demonstrate the superiority of TESS over BERT and RoBERTa, especially in small training sample settings.

We also prove the value of building social science-specific language models, as many texts encountered by social scientists are quite different from the generic text corpora used by computer scientists. This project shows that the path to studying pre-trained DNNs in the social science context may be fruitful. We encourage future researchers to explore whether certain pretraining tasks may serve social scientists better than the standard masked language modeling or next word prediction.

Since we do not include text from news outlets and Twitter data, we overlook a significant portion of the text that political scientists often use. It is unclear whether TESS can outperform BERT and RoBERTa on the news or Twitter data, as TESS may specialize in government documents at the expense of understanding other text types. For that reason, we recommend using TESS for processing text similar to our pretraining corpora while remaining cautious if adapting it to a new domain.

\printbibliography
\end{document}